\documentclass[sigconf]{acmart}
\AtBeginDocument{%
  }

\acmConference[]{}{}{}
  
\usepackage{graphicx}
\usepackage{float}
\usepackage[normalem]{ulem}

\usepackage{enumitem}
\usepackage[most]{tcolorbox}  

\renewcommand\footnotetextcopyrightpermission[1]{}
\begin{document}

\title{CORE-KG: An LLM-Driven Knowledge Graph Construction Framework for Human Smuggling Networks}

\thanks{Note: All example names in this paper are anonymized or fictitious to protect privacy. Full proper names and company's names are replaced with initials followed by a period (e.g., "R.").}

\author{Dipak Meher}
\affiliation{%
  \institution{George Mason University}
  \city{Fairfax}
  \country{USA}}
\email{dmeher@gmu.edu}

\author{Carlotta Domeniconi}
\affiliation{%
  \institution{George Mason University}
  \city{Fairfax}
  \country{USA}}
\email{cdomenic@gmu.edu}

\author{Guadalupe Correa-Cabrera}
\affiliation{%
  \institution{George Mason University}
  \city{Fairfax}
  \country{USA}}
\email{gcorreac@gmu.edu}

\begin{abstract}
Human smuggling networks are increasingly adaptive and difficult to analyze. Legal case documents offer valuable insights but are unstructured, lexically dense, and filled with ambiguous or shifting references—posing challenges for automated knowledge graph (KG) construction. Existing KG methods often rely on static templates and lack coreference resolution, while recent LLM-based approaches frequently produce noisy, fragmented graphs due to hallucinations, and duplicate nodes caused by a lack of guided extraction. We propose CORE-KG, a modular framework for building interpretable KGs from legal texts. It uses a two-step pipeline: (1) type-aware coreference resolution via sequential, structured LLM prompts, and (2) entity and relationship extraction using domain-guided instructions, built on an adapted GraphRAG framework. CORE-KG reduces node duplication by 33.28\%, and legal noise by 38.37\% compared to a GraphRAG-based baseline—resulting in cleaner and more coherent graph structures. These improvements make CORE-KG a strong foundation for analyzing complex criminal networks.
\end{abstract}

\keywords{Knowledge Graph Construction, Coreference Resolution, Human Smuggling Networks}


\maketitle

\section{Introduction}

Human smuggling has evolved into a complex and organized operation involving dynamic networks of actors, routes, vehicles, and intermediaries \cite{carrasco2025scapegoating}. In human smuggling networks, smugglers facilitate human mobility through illicit means, typically in exchange for a fee and within the context of restrictive immigration policies.
These networks exploit legal loopholes, adapt rapidly to enforcement policies, and often intersect with transnational criminal organizations. Understanding the structure and behavior of such illicit systems is essential for crafting effective policy, enhancing security, and preventing exploitation. Yet, much of the actionable intelligence about smuggling networks remains buried in unstructured legal texts—such as court documents, field reports, and case transcripts.

Despite growing interest from legal and social science communities, computational approaches for analyzing these documents remain underdeveloped. Entity references in unstructured legal text are frequently inconsistent—appearing as aliases, abbreviations, or role-based titles (e.g., “Officer R.” vs. “Defendant R.”)—which complicates coreference resolution, entity normalization, and downstream tasks like  extraction and knowledge graph construction.

Prior work has demonstrated the utility of knowledge graphs in legal investigations. For example, Mazepa et al.\cite{mazepa2022relationships} and Shi et al.\cite{shi2022knowledge} used rule-based and regular-expression methods to construct graphs from homicide and indictment cases. However, these systems rely on static templates and lack the flexibility to resolve aliasing or surface-level entity variation—making them less robust in handling complex, multi-entity narratives.

While LLMs have shown strong potential in general knowledge extraction~\cite{le2023large}, their application to constructing knowledge graphs from human smuggling case narratives remains underexplored. In particular, challenges such as resolving surface-level redundancy across semantically equivalent mentions within typed entities—especially in heterogeneous, multi-entity documents like smuggling cases—are still largely unaddressed~\cite{ji2020deep}. This represents a critical gap, as failure to consolidate such mentions leads to fragmented and redundant graph representations. Moreover, standard LLM-based extractors frequently misclassify or hallucinate entities and relations~\cite{zhang2024extract, huaman2020duplication}, introducing semantic noise that impedes downstream analysis.

To address these challenges, we introduce CORE-KG —a modular, prompt-driven LLM-based framework for constructing precise and interpretable knowledge graphs from legal case documents. CORE-KG comprises two key components: (1) a \textit{type-aware coreference resolution} module that consolidates semantically and contextually equivalent mentions within each entity type, and (2) a \textit{knowledge graph construction} module that uses structured prompts to extract entities and relationships. These prompts include domain-specific filtering instructions to suppress legal boilerplate, sequential type-wise extraction to minimize attention diffusion, and explicit type definitions to reduce classification ambiguity.

We evaluate CORE-KG on U.S. federal and state court cases related to human smuggling, drawn from publicly available legal proceedings. In the absence of annotated ground-truth graphs, we benchmark against a GraphRAG-based baseline \cite{edge2024local}, where both systems construct knowledge graphs from unstructured legal text using large language models applied over chunked inputs. On average, CORE-KG reduces node duplication by 33.28\%, and legal noise by 38.37\% in the resulting knowledge graphs. In a representative case study, CORE-KG identifies key actors, critical transit routes, transport vehicles, involved organizations, smuggled items, and communication methods—revealing structural insights essential for downstream analysis. \footnote{Our code is available at \url{https://github.com/dipakmeher/CoreKG-HumanSmuggling}.}

\vspace{1mm}
\noindent To summarize, our key contributions are:
\begin{itemize}
    \item We present CORE-KG, a modular, LLM-based framework that integrates type-aware coreference resolution and structured, domain-specific prompting to construct knowledge graphs from legal case documents on human smuggling.
    
    \item We introduce a prompt-based coreference resolution module that resolves semantically and contextually equivalent mentions across diverse entity types, enhancing consistency in structured knowledge extraction from complex narratives.
    
    \item CORE-KG outperforms a GraphRAG baseline, achieving a 33.28\% reduction in node duplication, and 38.37\% reduction in legal noise.
\end{itemize}

\section{Related Work}

Knowledge graphs are powerful tools for transforming unstructured text into structured representations that facilitate reasoning, retrieval, and analysis~\cite{edge2024local, zhong2023comprehensive}. Recently, they have been applied to various tasks across domains such as education~\cite{chen2018knowedu}, life sciences~\cite{callahan2024open}, and construction safety management~\cite{fang2020knowledge}. However, if not constructed carefully, they often suffer from issues such as duplicate nodes and fragmented structures, which significantly reduce their utility~\cite{huaman2020duplication}.

\subsection{Knowledge Graph Construction}

Traditionally, knowledge graphs have been built by first identifying important entities and their relationships from text, and then connecting them step by step. Entity extraction techniques include rule-based and dictionary-based methods~\cite{sun2018overview}, statistical and machine learning-based models~\cite{sun2018overview}, and open-domain or domain-specific extraction approaches~\cite{sun2018overview, liu2011recognizing, ling2012fine}. For relation extraction, researchers have employed syntactic, lexical, and semantic modeling~\cite{kambhatla2004combining}, ontology-based frameworks such as HowNet~\cite{liu2007implementation}, and semi-supervised learning techniques~\cite{carlson2010coupled}. While useful in specific domains, these methods often depend on hand-crafted extraction patterns and aim to preserve alignment between the graph and the source text.

More recently, LLM-based frameworks have emerged as powerful tools for constructing knowledge graphs in a prompt-driven manner~\cite{zhang2024extract, vizcarra2024representing}. Kommineni et al.~\cite{kommineni2024human} proposed a semi-automated pipeline that integrates competency question generation, ontology design, and RAG-based triple extraction from scholarly texts. Zhang et al.~\cite{zhang2024extract} introduced a modular framework that performs open information extraction, schema definition, and canonicalization to reduce redundancy and improve consistency. While these approaches minimize manual effort, they often assume clean, unambiguous input and do not account for challenges like reference ambiguity or entity aliasing. This becomes particularly problematic in domains such as law, where entities frequently shift between aliases, roles, and pronouns.

 To mitigate such challenges, coreference resolution plays a key role in maintaining graph coherence. Wang et al.~\cite{wang2020coreference}, for instance, showed that even infrequent pronoun mentions can significantly affect the completeness of educational-domain knowledge graphs. This highlights the importance of integrating coreference mechanisms directly into the graph construction pipeline—especially when entity references are inconsistent or implicit.

Several efforts have also explored knowledge graph construction in criminal domains. Mazepa et al.~\cite{mazepa2022relationships} constructed a knowledge graph for homicide investigations using rule-based NLP pipelines and CoreNLP components. Shi et al.~\cite{shi2022knowledge} developed a Neo4j-based graph for job-related crime indictments using regular-expression-based entity and relationship extraction. While both systems demonstrate the utility of knowledge graphs in legal and investigative contexts, they rely heavily on static templates and lack both coreference handling and modular prompting strategies—limiting their robustness in complex, ambiguous narratives.

Our work addresses this gap by introducing a prompt-driven, entity-type-specific coreference resolution strategy tailored for the modeling of criminal networks. This design improves semantic precision and node coherence in the resulting  knowledge graphs, particularly in cases involving complex inter-entity relationships such as criminal networks, migration routes, and procedural actors.

\subsection{Coreference Resolution}
Several methods have been proposed to address the problem of duplicate nodes and fragmented structures through coreference resolution~\cite{liu2023brief, lata2022mention}. Pogorilyy et al.~\cite{pogorilyy2019coreference} and Wu et al.~\cite{wu2017deep} each proposed coreference resolution methods based on convolutional neural networks (CNNs), demonstrating improvements in modeling semantic and syntactic patterns in text.

Recent advances in large language models (LLMs) have brought new momentum to coreference resolution, particularly in low-resource settings where annotated data is scarce. Prompt-based approaches have shown that even with minimal supervision, LLMs can perform competitively by leveraging few-shot, zero-shot, or chain-of-thought strategies~\cite{das2024co, gan2024assessing}. These methods typically operate on datasets involving either a single entity type—often restricted to human mentions—or on general narrative text where entity-type distinctions are not explicitly modeled~\cite{das2024co, tran2025coreference}. While such setups enable LLMs to maintain focused attention and achieve high performance, they do not reflect the complexity of domain-specific texts where entities span diverse types and are referenced inconsistently across roles. As a result, the ability of LLMs to generalize across entity types and resolve fine-grained, role-shifting references remains largely untested.

Some progress has also been made in coreference resolution for legal texts. The work by Jia et al.~\cite{ji2020deep}  introduces a neural network-based model using ELMo, BiLSTM, and GCNs to resolve speaker-based coreference in court records. To the best of our knowledge, there is no prior research that investigates coreference resolution in the legal domain using large language models.
Recent work has shown that instruction-tuned LLMs can achieve competitive performance in coreference resolution \cite{le2023large}, but these evaluations are limited to standard narrative data and do not address the challenges posed by multi-entity legal documents.



\section{Method}

\subsection{Method Overview}

Our proposed pipeline is designed for modular knowledge graph construction (KG) from narrative-rich documents. While applicable across domains, we demonstrate its effectiveness on legal case files, with a focus on identifying actors, events, and entities involved in human smuggling networks. Specifically, we extract seven types of entities: \textit{Person}, \textit{Location}, \textit{Organization}, \textit{Route}, \textit{Means of Transportation}, \textit{Means of Communication}, and \textit{Smuggled Items}. The pipeline consists of the following three major components:

\begin{itemize}
    \item \textbf{Coreference Resolution:} We apply entity-type-aware coreference resolution to unify both surface and contextually grounded references to the same real-world entity (e.g., “Y.”, “A.Y.”, “Defendant”). This improves graph coherence and reduces redundancy, while preserving distinctions when context indicates separate referents.

    \item \textbf{Prompt Optimization for Entity and Relation Extraction:} We design a contextualized prompt that combines established prompting principles with domain-specific strategies for accurate entity and relation extraction. Although the framework is broadly applicable, we demonstrate its effectiveness for the task of building a KG for human smuggling networks. The prompt integrates sequential entity-type extraction, in-prompt type definitions, and explicit filtering instructions tailored to the downstream task of criminal network analysis. These components are organized in a chain-of-thought structure to reduce attention drift, improve type disambiguation, and eliminate irrelevant or noisy extractions.

    \item \textbf{KG Construction:} Entities and relationships extracted by a large language model (LLM), guided by a tailored prompt, are assembled into a structured knowledge graph using the GraphRAG knowledge graph construction module.
\end{itemize}

Since legal documents often contain procedural or statutory sections that are not central to the case narrative, we extract the \textit{Opinion section} as input to our pipeline. This section typically contains the factual context and actor interactions most relevant to human smuggling cases. We apply coreference resolution to unify entity mentions, and then process the resolved text using the KG construction module to extract structured entities and relationships and build the final graph.

\subsection{Coreference Resolution}

Legal documents frequently refer to the same real-world entity using different surface forms throughout the text. For instance, a defendant introduced as “A.Y.” may later be referred to as “Y.”, “the defendant,” or functionally as “the driver.” While this example concerns a \texttt{Person} entity type, similar patterns occur across other types as well. If these coreferent mentions are not resolved prior to the construction of the KG, they result in multiple disconnected nodes representing the same entity, fragmenting relationships and undermining graph interpretability. Such fragmentation dilutes the structural coherence and weakens the utility of the graph for downstream analysis.

\subsubsection{Coreference Resolution Module}

To address entity fragmentation, our Coreference Resolution module leverages the contextual reasoning capabilities of large language models (LLMs) \cite{wang2023can,havrilla2024glore} to unify different surface forms and contextually grounded references to the same real-world entity into a single, consistent representation. For example, references like "Y.", and "the defendant," are resolved to one canonical form. This unification ensures that all relationships tied to that entity are aggregated under a single node in the KG, improving both structural consistency (avoiding redundant nodes) and analytical clarity (making entity relationships easier to interpret).

The module operates over the Opinion section of each legal case and performs coreference resolution separately for each entity type with the help of an LLM. LLMs, trained on large and diverse text corpora, excel at understanding context, reasoning over long spans of text, and linking semantically related phrases \cite{an2024make, meherunderstanding, ding2024longrope}. However, they perform best when their attention is guided toward a specific objective. When multiple types are resolved simultaneously, the model’s attention becomes diffusely distributed across semantically unrelated categories (e.g., persons, route, organizations, etc.) \cite{abdelnabi2024you}. Since LLMs rely on self-attention mechanisms—designed to dynamically assign importance to different spans of text—this broad distribution leads to diluted focus, making it harder for the model to form accurate coreference chains, i.e., correctly linking all references to the same entity.

From a representation learning perspective, mixing multiple semantic types in a single pass increases the risk of type drift (where an entity is assigned the wrong category) and feature entanglement (where features from different types interfere with one another) \cite{zhou2023universalner}. These errors can propagate into downstream tasks like entity linking or graph construction, leading to noisy or incomplete KGs. This issue is exacerbated when surface forms are ambiguous or overlap semantically. For example, the phrase “The Camp” might refer to a temporary location used for migrant holding (a Location) but could also be misclassified as an Organization, especially when entity boundaries are vague or context is minimal. Similarly, “the van” might refer to a vehicle used for transport (Means of Transportation), but in some contexts, it may be interpreted as the location where migrants are held (Location). Such ambiguity in surface forms often results in incorrect type assignments and fragmented entity clusters in the graph.

To mitigate these risks, we adopt a type-wise sequential resolution strategy. The model resolves one entity type at a time, which reduces cross-type interference and allows more accurate consolidation of coreferences. 
The pipeline operates as follows:
\begin{enumerate}
    \item The filtered opinion text is first passed to the LLM along with a prompt tailored to resolve \texttt{Person} entities. Mentions like "Y.", "the defendant", and "the driver" are linked and replaced with their canonical reference.
    \item The output from this step is then used in the next stage, where a \texttt{Location} prompt resolves place-based mentions like "Laredo" and "Laredo Texas."
    \item The process continues in sequence for the remaining entity types: \texttt{Routes}, \texttt{Organization}, \texttt{Means of Transportation}, \texttt{Means of Communication}, and \texttt{Smuggled Items}.
\end{enumerate}

This targeted approach improves semantic precision (correct type assignment) and structural fidelity (clear, non-overlapping entity representation) in the resulting graph. With input from a subject matter expert, we limit the resolution process to seven entity types that are most relevant to analyzing smuggling networks: Person, Location, Routes, Means of Transportation, Means of Communication, Organization, and Smuggled Items. By focusing only on these categories, the model avoids overgeneralization and stays aligned with the goals of criminal network modeling.

\subsubsection{Coreference Resolution Prompt Design}
\label{sec:coref_resolution_prompt_design}
Since coreference resolution is performed separately for each entity type, we design a dedicated prompt for every entity type of interest. Each prompt is paired with the input text to guide the language model in resolving coreferences for that specific category only.

Extensive prompt optimization is performed independently for each entity type to maximize the quality and completeness of coreference resolution. The prompts are designed through iterative refinement and evaluation, ensuring that they generalize well across diverse legal cases. Particular attention is given to capturing the wide variability in surface forms within each category. For example, \texttt{Person} entities may appear as full names, shortened names, aliases, and roles (e.g., \texttt{Defendant}, \texttt{Driver}). \texttt{Location} entities may be expressed with varying granularity (e.g., \texttt{Laredo}, \texttt{"Laredo, Texas"}, while \texttt{Routes} can be denoted through formal identifiers or colloquial expressions (e.g., \texttt{Interstate 35}, \texttt{I-35}). Similarly, \texttt{Organizations} may include government bodies, company names, or abbreviations, and \texttt{Means of Transportation} often blend object and ownership references (e.g., \texttt{Trailer}, \texttt{Tractor-Trailer}). 
In designing these prompts, we intentionally avoided overfitting to any fixed phrase list and instead focused on creating entity-type-specific guidelines and few-shot examples that reflect the semantic diversity and ambiguity present in real legal documents. As a result, the prompts are intended to support generalization to unseen or unexpected surface forms. In Section \ref{sec:rq1_coref} we discuss several qualitative examples where the model correctly resolved such references, including collective terms like “relatives of the aliens”.

To maintain consistency and precision across these varied patterns, each prompt follows a structured format containing the below key components:

\begin{itemize}
    \item \textbf{Persona definition}: The LLM is assigned the role of a highly accurate and rule-following coreference resolution system.
    \item \textbf{Clear task description}: The prompt clearly states that the LLM must resolve all coreferences without summarizing, rephrasing, or altering the input text in any way.
    \item \textbf{Contextual information}: The prompt explains the downstream use of the resolved text, such as for building knowledge graphs to analyze human smuggling networks.
    \item \textbf{Entity-type-specific resolution rules}: Tailored resolution guidelines are provided for each entity type based on their linguistic patterns. For instance, the rules for the \texttt{Person} entity type address variations such as full names, abbreviations, last names, aliases, and role-based references (e.g., ``the driver'', ``the defendant'').
    \item \textbf{Few-shot examples}: Illustrative examples are provided to show the model how to correctly resolve coreferences in different realistic scenarios.
\end{itemize}

An example of the coreference resolution prompt used for the \texttt{Person} entity type is provided in Appendix Figure~\ref{fig:coref_prompt_person}. This structured and detailed prompt design ensures that the model receives precise guidance, thereby improving the accuracy and consistency of coreference resolution across diverse legal documents.

\subsection{Entity-Relationship Extraction and KG Construction}

The coreference-resolved Opinion section serves as input to the Knowledge Graph Construction (KGC) module of the GraphRAG framework \cite{edge2024local}. GraphRAG is a modular retrieval-augmented generation (RAG) system that combines structured knowledge graph construction with neural retrieval and language model-based generation. It has gained traction for its ability to mitigate hallucination and improve factual grounding in downstream tasks by anchoring generation to a graph-based representation of the input text.

GraphRAG consists of two core components: (1) a KGC module that extracts entities and relationships to form a structured graph, and (2) a retrieval-augmented generation module that leverages this graph as an index for response generation. This work concentrates specifically on the KGC component, which we extend and tailor to the processing of legal case files for human smuggling network analysis.

Each input document is divided into overlapping chunks of 300 tokens, which are passed to the LLM along with an extraction prompt optimized for the legal domain of human smuggling cases. The model outputs entity–relationship triples, which are aggregated across chunks. GraphRAG performs basic post-processing—merging entities based on exact string and type matches—and assembles the graph using the \texttt{NetworkX} library. Final outputs are serialized in both GraphML and Parquet formats to support further analysis and visualization via tools like Gephi or the GraphRAG visualizer.


\subsubsection{Prompt Tuning for Entity and Relationship Extraction} 
\label{sec:prompt_tuning}

The prompt for entity and relationship extraction follows the same principled design strategies outlined in Section~\ref{sec:coref_resolution_prompt_design}. However, unlike the coreference resolution module—where prompts are applied sequentially for each entity type—GraphRAG’s  KGC component requires a single unified prompt to extract all relevant entity types and their relationships in one pass. 

To accommodate this, the prompt is carefully designed to balance breadth and precision across the seven targeted types, enabling the LLM to generate structured triples with minimal noise and high task relevance. It incorporates the same components as the coreference prompts—namely, persona definition, task description, contextual framing, extraction steps, output format specification, and few-shot examples—as detailed in Section~\ref{sec:coref_resolution_prompt_design}. An example of the final prompt used for entity and relationship extraction is shown in Appendix Figure~\ref{fig:llm_prompt}.


To further improve the quality of entity and relationship extraction for human smuggling network analysis, we introduce several additional modifications to the prompt design, as described below.

\paragraph{Sequential Entity Extraction to Reduce Attention Distribution}  
GraphRAG’s default prompt extracts all entity types jointly, which can lead to attention spread across multiple categories—causing the model to miss entities, assign incorrect types, or overlook fine-grained distinctions \cite{abdelnabi2025get}. These issues are especially pronounced in legal texts, where diverse entity types frequently co-occur within complex, narrative structures.

To mitigate this, we introduced a strict ordering mechanism within the prompt: the model is explicitly instructed to first extract entities of type \textit{Person}, followed by \textit{Location}, \textit{Routes}, \textit{Organization}, and so on. Only after completing entity extraction in the specified sequence does the model proceed to identify relationships among the extracted entities. By enforcing sequential extraction by entity type, we significantly reduce competition for attention within the model, enabling it to capture all mentions of one entity category before shifting focus to the next—thereby improving both the precision of extracted entities and the completeness of relationships, which directly contributes to cleaner and more reliable knowledge graph construction.

\paragraph{Filtering High-Frequency Irrelevant Entities}
Another significant challenge arises from the presence of high-frequency but contextually irrelevant entities within legal documents, particularly government-related entities such as courts, juries, appeals, and judicial offices. These entities are typically classified under the \textit{Organization} type during extraction. However, if included in the final knowledge graph, these non-critical entities tend to dominate the node distribution and inflate relationship statistics due to their frequent mentions in legal proceedings, which introduces noise and obscures the true patterns of interaction among the smuggling participants.

To address this issue during extraction itself, we introduced an explicit filtering instruction within the prompt. After completing the entity and relationship extraction steps, the LLM is guided to identify all government-related nodes based on predefined criteria and remove them before producing the final output. By integrating this filtering at the extraction stage, we eliminate the need for separate post-processing to clean the graph. This early-stage filtering improves the overall quality, clarity, and analytical usefulness of the resulting knowledge graphs by ensuring that only domain-relevant entities are retained.

\paragraph{Entity Type Definitions to Mitigate Overgeneralization Bias}
Large language models (LLMs), although trained on massive and diverse datasets and capable of strong reasoning and generalization, sometimes suffer from \textit{overgeneralization bias} — a phenomenon where models incorrectly classify entities based on statistical patterns or common co-occurrences observed during pretraining, rather than the specific context of the input \cite{peters2025generalization, dai2024bias}. For example, if certain words frequently appear near geographic locations in the training data, the model might incorrectly classify unrelated terms such as event names, organizational units, or descriptors as \textit{Location} entities during extraction, even if they are not true locations within the legal document.

Such misclassifications are especially problematic in legal narratives, where multiple specialized terms coexist and subtle contextual cues determine the correct entity type. To mitigate this, our prompt design includes clear, concise definitions for each of the seven targeted entity types. These definitions are embedded directly within the prompt instructions, specifying what qualifies as a \textit{Person}, \textit{Location}, \textit{Route}, \textit{Organization}, and so on, with illustrative examples provided where needed. By incorporating explicit type definitions into the prompt, we reduce the model’s reliance on vague pretraining associations and guide it toward making context-sensitive classification decisions, thereby improving extraction precision, reducing type confusion, and enhancing the overall consistency and reliability of the resulting knowledge graph.



\section{Experimental Setup}

We aim to answer the following research questions:
\begin{itemize}
   \item \textbf{RQ1:} Does incorporating prompt-based coreference resolution reduce node duplication in knowledge graphs constructed from human smuggling case documents?

    \item \textbf{RQ2:} Does CORE-KG—our fully adapted pipeline incorporating coreference resolution, prompt tuning, and legal-specific filtering—generate more relevant and concise knowledge graphs than the GraphRAG-based baseline?
\end{itemize}

\subsection{Dataset}
To evaluate our system we use judicial cases accessed via the academic search engine Nexis Uni, which is available to us through the library of our Institution. We focus on cases related to human smuggling networks. These documents are drawn from federal and state court proceedings filed between 1994 and 2024, providing a diverse and representative sample of criminal smuggling activities across jurisdictions. All information referenced in these cases is part of the public domain.

However, legal case files often include extensive procedural, statutory, and administrative content, which introduces noise and redundancy when constructing knowledge graphs that aim to capture the structure of criminal networks. To mitigate this, we apply targeted preprocessing to extract only the section entitled ``Opinion" from each case. This section contains the main factual narrative, describing the people involved, the routes used, the items transported, and the sequence of events. These elements are directly relevant for building meaningful and interpretable knowledge graphs. All methods compared in our experiments are provided with the same input text from the Opinion section, hereafter referred to as the input document.

In our preliminary experiments we have randomly selected a sample of 20 cases (retrieved from the Nexis Uni database using the query: ``{\textit{human smuggling OR alien smuggling}"), in which the Opinion section contains approximately 2000 words. Although a larger sample size is needed for comprehensive analysis, the selected cases sufficiently reflect the narrative complexity and are suitable for a preliminary assessment of the system’s capacity to extract key entities and identify meaningful relationships and interactions.


\subsection{Implementation}

We employ the LLaMA 3.3 70B model for both coreference resolution and knowledge graph construction. To ensure reproducibility and minimize stochastic variation during inference, the temperature parameter is set to zero. The model is served locally using the Ollama framework, allowing us to operate entirely within an open-source environment without relying on commercial or closed-source APIs. All experiments were conducted on an NVIDIA A100 GPU with 80GB of memory.

Our codebase is implemented in Python 3.12. For GraphRAG \cite{edge2024local}, we follow the setup instructions provided in the official GitHub repository and use version 0.3.2. We configured GraphRAG to use 300-token chunks and specified the nomic-embed-text embedding model; however, embeddings were not utilized, as the knowledge graph construction process does not rely on them.

\subsection{Baselines and Experimental Design}

Given the lack of prior work on structured knowledge graph construction from legal texts—especially for criminal network analysis such as human smuggling—we evaluate our system against a baseline derived from the GraphRAG framework \cite{edge2024local}, which is widely adopted for knowledge-intensive tasks. We compare two system variants:

\begin{itemize}
    \item \textbf{GraphRAG (Baseline):} A minimally adapted version of the default GraphRAG prompt \cite{edge2024local}. It includes the addition of the seven targeted entity types and a few-shot example to guide extraction, but retains the original model behavior, output format, and processing flow. 
    
    \item \textbf{CORE-KG (Ours):} A fully adapted pipeline that integrates several domain-specific enhancements. These include (1) prompt-based coreference resolution, (2) sequential entity extraction to reduce attention dilution, (3) in-prompt entity type definitions and negative examples to guide classification, and (4) legal-specific filtering instructions to suppress noisy or irrelevant entities.
\end{itemize}

Since no annotated ground truth exists for structured knowledge graphs in this domain, we adopt both a qualitative and quantitative evaluation framework. Both systems are applied to the same 20 input documents. 
The resulting graphs are  evaluated as follows:

\begin{itemize}
    \item \textbf{Node Duplication:} We compute the number of redundant nodes representing the same real-world entity (e.g., \texttt{Y.}, \texttt{A.Y.}, \texttt{Defendant}). This process is partially automated using intra-type fuzzy matching (see Section \ref{sec:rq1_coref}).
    
    \item \textbf{Noise Reduction:} We document the inclusion of generic legal boilerplate or irrelevant procedural terms (e.g., \texttt{Judicial Proceedings}, \texttt{Appeal Process}, \texttt{Court}) that do not aid downstream smuggling network analysis, and assess their suppression across systems.

    
    \item \textbf{Type Assignment Reliability:} We qualitatively examine common misclassifications of entities (e.g., labeling ``The Safehouse'' as an organization instead of a location) and evaluate the impact of prompt-level refinements in reducing such errors.
\end{itemize}

This evaluation protocol enables a focused comparison between the baseline and CORE-KG outputs, isolating the effects of coreference resolution and prompt engineering in building domain-aligned, analytically useful knowledge graphs.

\section{Results}

\subsection{RQ1: Impact of Coreference Resolution on Node}
\label{sec:rq1_coref}

To evaluate the impact of coreference resolution, we analyze the extent of node duplication in both the baseline system and our CORE-KG pipeline. Duplicate nodes are defined as semantically equivalent entities that appear multiple times in the graph under different surface forms (e.g., \texttt{Y.}, \texttt{A.Y.}).

\paragraph{Duplicate Node Detection} Duplicate node detection is performed through a two-stage process. First, we apply fuzzy string matching to the extracted entities using the \texttt{partial\_ratio} function from the \texttt{RapidFuzz} library, retaining all intra-type entity pairs with similarity scores at or above a 75\% threshold. These pairs are used to construct an undirected similarity graph, from which clusters are extracted as connected components—each assumed to represent a single real-world entity. For example, mentions such as \texttt{white pickup truck}, \texttt{stolen white pickup truck}, and \texttt{white older Ford pickup truck} would be grouped into the same cluster. In the second stage, a manual review is conducted by a subject matter expert to correct false positives—i.e., cases where distinct entities were incorrectly grouped together.

The final node duplication count is computed as $\sum_{C_i} (|C_i| - 1)$, where $C_i$ denotes a cluster of mentions referring to the same entity. To allow for comparability across graphs of varying sizes, we report the \textit{node duplication rate} as the number of redundant nodes divided by the total number of nodes in the graph. This value represents the proportion of all entity nodes that are considered duplicates. \textit{Absolute Drop} is computed as the direct difference between the baseline and CORE-KG scores. \textit{Relative Improvement} is calculated as the percentage reduction relative to the baseline: $(\text{Baseline} - \text{CORE-KG}) / \text{Baseline} \times 100$.

\paragraph{Noise Detection}
To enhance the quality and interpretability of the generated knowledge graphs, we conduct a systematic manual validation of the extracted entities in each case. This validation, performed by a domain expert, involves identifying legal boilerplate or procedural terms (e.g., \texttt{Court}, \texttt{Appeal}, \texttt{Judicial Proceedings}) that do not contribute to the structural understanding of the smuggling network. Although frequent in legal texts, such entities are deemed semantically irrelevant for downstream analysis. We quantify noise using the \textit{noise rate}, defined as the number of non-informative nodes divided by the total number of nodes in the graph, multiplied by 100.

\begin{table}[h]
\centering
\resizebox{\columnwidth}{!}{%
\begin{tabular}{@{}lcccc@{}}
\toprule
\textbf{Metric} & \textbf{Baseline (\%)} & \textbf{CORE-KG (\%)} & \textbf{Absolute Drop (\%)} & \textbf{Relative Improvement (\%)} \\
\midrule
Node Duplication Rate         & 30.38 & 20.27 & 10.11 & 33.28 \\
Noise Rate                    & 27.41 & 16.89 & 10.52 & 38.37 \\
\bottomrule
\end{tabular}
}
\caption{Comparison of CORE-KG and the baseline across three key evaluation metrics aggregated over 20 legal cases.}
\label{tab:corekg_full_metrics}
\end{table}

Table~\ref{tab:corekg_full_metrics} presents the average duplication statistics across 20 legal case graphs, comparing the graphs generated by the baseline GraphRAG system with those produced by our CORE-KG pipeline. We observe a substantial reduction in node duplication, driven by the improvements introduced through entity-type-aware prompt-based coreference resolution. In the baseline, 30.38\% of all nodes in the graph were redundant mentions of entities already captured by another node in their cluster. CORE-KG reduces this to 20.27\%, yielding a 33.28\% relative improvement. These results underscore the effectiveness of our pipeline in resolving coreference, collapsing redundant structures, and producing cleaner, more coherent knowledge graphs. Detailed case-wise comparisons for all three metrics across the 20 legal cases are presented in Appendix~\ref{sec:appendix_percase_graphs}, showing consistent performance improvements of CORE-KG over the baseline.

\begin{table*}[t]
\centering
\small
\begin{tabular}{|p{3.2cm}|p{5.8cm}|p{6.8cm}|}
\hline
\textbf{Issue Type} & \textbf{Observed in Baseline} & \textbf{CORE-KG Behavior and Reason for Improvement} \\
\hline
Duplicate Entities & Multiple variants like \texttt{Y.}, \texttt{A.Y.}; \texttt{J.C.D.A}, \texttt{Agent D.A.}; \texttt{Laredo}, \texttt{Laredo Texas}; \texttt{I-35}, \texttt{Interstate 35}, \texttt{Route from Laredo to Dallas}; \texttt{B.}, \texttt{Y.B.}. & Consolidated into unified nodes through entity-type-specific coreference resolution: \texttt{A.Y.}, \texttt{J.C.D.A}, \texttt{Laredo Texas}, \texttt{Interstate 35}, \texttt{Y.B.}. \\
\hline
Generic Legal Entities &Entities like \texttt{court}, \texttt{district court}, \texttt{court of appeals}, \texttt{state court}, \texttt{appeal}, \texttt{appeal process}, \texttt{judgment of acquittal}, \texttt{motion for judgment of acquittal}, and \texttt{plain error standard} clutter the graph and lack task relevance. & Removed through in-prompt filtering and reasoning-based instructions designed to retain only entities relevant to downstream analysis of smuggling networks. \\
\hline
Entity Type Misclassification
& Examples include \texttt{Court Case} and \texttt{Federal Proceedings} misclassified as \texttt{Location}; \texttt{United States Magistrate Judge} mislabeled as \texttt{Organization} instead of \texttt{Person}; \texttt{Jamco} tagged as \texttt{Location} rather than \texttt{Organization}; and abstract terms like \texttt{Verdict} or \texttt{Ineffective Assistance} incorrectly extracted as \texttt{Smuggled Items}.
& CORE-KG reduces such misclassifications through sequential, type-specific extraction and explicit in-prompt definitions that help the LLM disambiguate and align entity roles correctly. \\
\hline
\end{tabular}
\caption{Qualitative Comparison of Prompt Effectiveness: CORE-KG vs. Baseline on \textit{a representative legal case}}
\label{tab:rq2_qualitative_comparison}
\end{table*}

\subsubsection{CORE-KG’s Context-Aware and Generalization Capability in Coreference Resolution}

While conducting error analysis on unresolved duplicates, we found that not all apparent misses were true failures. For example, in Case 6, phrases like “undocumented alien,” “illegal trafficking in aliens,” and “smuggling aliens” were not merged by CORE-KG, despite their surface similarity. Closer inspection revealed that these referred to distinct concepts: a specific individual in custody, a general criminal activity, and broader trafficking patterns along a route. By preserving these distinctions, CORE-KG demonstrated sensitivity to discourse context and avoided spurious generalization.


In Case 13, mentions such as “relatives,” “relatives of the aliens,” and “family members”—all referring to individuals in New York connected to C.E.C.M.—were initially extracted as separate nodes by the baseline. CORE-KG correctly unified them into a single node labeled \texttt{C.E.C.M's Relatives}, capturing indirect, collective references absent from the prompt.

These examples provide strong qualitative evidence that CORE-KG does not indiscriminately cluster surface-similar mentions. Instead, it applies contextual judgment—guided by structured prompts and LLM inference—to balance generalization with precision across noisy, heterogeneous legal narratives.




\subsubsection{Error Analysis: Missed Coreference Resolutions}
That said, we did observe a few genuine misses. For example, in Case 15, the mentions “United States–Mexican border” and “border” were not clustered, and in Case 10, “United States government” and “United States” were treated as separate entities. We acknowledge these as true missed coreference cases. A likely reason is that legal documents often contain high lexical density and institutional specificity, where certain entity mentions—although contextually coreferent—appear in subtly distinct syntactic and semantic frames. In such contexts, the LLM may treat “border” as a general geographic location and “United States–Mexican border” as a formal, compound entity. Similarly, “United States” may refer either to the country as a location or the government as an actor, depending on local context. Without stronger global discourse tracking or reinforcement of named entity role alignment, these cases may fall outside the model’s resolution threshold.

These observations suggest opportunities for improving prompt sensitivity to geopolitical and institutionally anchored entities, particularly in domains with formal legal discourse.


\subsection{RQ2: Does CORE-KG produce more relevant and concise knowledge graphs compared to the baseline?}

In this section, we qualitatively assess the resulting knowledge graphs by evaluating the impact of prompt design across three dimensions: structural gain, reduction of noisy legal entity extractions, and resolution of entity type misclassifications.

\subsubsection{Structural Gains from Prompt-Guided Graph Construction}

To assess the structural benefits of CORE-KG, we compare its output against the baseline GraphRAG system using a representative case 2. Appendix Figures~\ref{fig:corekg_single_graph} and~\ref{fig:grag_single_graph} show the resulting knowledge graphs from both systems. In these visualizations, relationship labels are omitted for clarity, and node size reflects the number of connected relationships, serving as a proxy for centrality.

In the baseline graph, the primary actor—\texttt{A.Y.}—is fragmented across two separate nodes: \texttt{Y.}, and \texttt{A.Y.}. These nodes are only indirectly linked, resulting in a disjointed structure that inflates graph complexity and weakens narrative continuity. Notably, the most central node in the baseline graph is \texttt{Y.}, despite the existence of multiple redundant mentions referring to the same individual. In contrast, CORE-KG successfully consolidates all references into a single, canonical \texttt{A.Y.} node, producing a cleaner and semantically faithful representation of the actor’s role.

Further structural improvements are evident in the connectivity of key locations. In the baseline graph, \texttt{Laredo, Texas}—a critical geographic location in the smuggling operation is disconnected, illustrating how legal noise and lack of guidance in extraction can obscure essential entities. In CORE-KG, however, \texttt{Laredo, Texas} is clearly connected to relevant entities, including \texttt{A.Y.}, \texttt{Interstate 35}, \texttt{Illegal Aliens}, and \texttt{Agent J.C.D.A.}, reflecting a more complete and meaningful network of interactions. 

These improvements illustrate how CORE-KG’s guided extraction strategy produces more coherent, focused, and narratively aligned graphs from complex legal texts.
\label{sec:result_entity_prompt_impact}


\subsubsection{Reducing Legal Noise Through Prompt-Guided Filtering}

Since the input text is drawn from legal opinions, the baseline graph frequently includes noisy legal terms that are common in such documents but irrelevant to smuggling network analysis. In the qualitative example from Case 2, the baseline system extracted 28 such entities—including \texttt{state court}, \texttt{district court}, and \texttt{court of appeals}—which dilute the semantic clarity of the graph, as visualized in Figure~\ref{fig:noise_reduction} (Appendix). These nodes introduce noise and obscure the core smuggling narrative. Additional examples of such irrelevant legal entities are provided in Table~\ref{tab:rq2_qualitative_comparison}.

A key limitation of the baseline system is that it prompts the LLM to extract all seven entity types simultaneously, which leads to attention dilution and type confusion. CORE-KG addresses this through two key improvements. First, it includes explicit filtering instructions that direct the model to ignore procedural and legal boilerplate terms. Second, it employs in-prompt sequential extraction, allowing the model to focus on one entity type at a time in a fixed order. This structured design significantly improves extraction accuracy. The prompt used for extraction is shown in Appendix Figure~\ref{fig:llm_prompt}.

In the same legal case, CORE-KG extracted no legal noise entities, yielding a clean and focused graph. Quantitatively, CORE-KG reduces the overall legal noise rate from 27.41\% to 16.89\%, an absolute improvement of 10.52\% points ( see Table~\ref{tab:corekg_full_metrics}). This demonstrates the model’s improved ability to focus on domain-relevant entities and avoid cluttering the graph with noisy nodes.


\subsubsection{Impact of Prompt Design on Type Misclassification}

Another critical improvement observed in CORE-KG pertains to entity type classification. In the baseline system, the language model often misclassifies entities due to vague type boundaries and insufficient contextual guidance. For example, \texttt{United States Magistrate Judge} is incorrectly labeled as an \texttt{Organization} instead of a \texttt{Person}, and \texttt{South Laredo} is tagged as a \texttt{Route} rather than a \texttt{Location}. In the representative case, the model further misclassifies legal terms such as \texttt{EVIDENCE} and \texttt{Verdict} as \texttt{Smuggled Items}, and \texttt{APPEAL PROCESS} as a \texttt{Route}. Additionally, entities like \texttt{J.} are incorrectly extracted as \texttt{Location} instead of their correct type, \texttt{Organization}. These errors are compounded by the baseline's tendency to assign types to legal boilerplate terms that should not have been extracted in the first place. Such misclassifications distort the semantic structure of the graph and hinder downstream analysis of actor–resource–event dynamics central to understanding smuggling networks.

CORE-KG substantially reduces these errors through two key prompt-level interventions. First, it employs sequential type-wise extraction, instructing the model to focus on one entity type at a time. This design minimizes cross-type confusion, especially in cases with ambiguous lexical cues. Second, the prompt provides explicit, context-sensitive definitions for each entity type, clearly distinguishing, for example, between actionable smuggled items and procedural legal terminology. These refinements constrain the model’s output space and promote more accurate and schema-aligned entity typing.

In the representative case, CORE-KG not only avoids extracting out-of-scope legal concepts but also correctly resolves \texttt{J.} to its full form, \texttt{J.I. Inc.}, and classifies it as an \texttt{Organization}—a classification consistent with its semantic role in the document. These improvements collectively enhance the accuracy, interpretability, and task relevance of the resulting knowledge graph.


\section{Conclusion}
In this work, we presented CORE-KG, a modular LLM-driven framework for constructing interpretable knowledge graphs from legal case documents describing human smuggling networks. By integrating type-aware prompt-based coreference resolution and domain-specific prompt design, CORE-KG addresses key challenges such as duplicate nodes, noisy legal entities, and entity misclassification—common issues in both rule-based and general-purpose LLM-based systems. Through evaluation on real-world smuggling cases, CORE-KG demonstrated significant improvements over a GraphRAG-based baseline, achieving a 33.28\% reduction in node duplication, and 38.37\% reduction in legal noise. These gains result in cleaner, more coherent graphs that better support downstream analysis of criminal networks. Our results underscore the importance of structured prompting, entity-type separation, and legal-domain adaptation for improving the quality of LLM-generated knowledge graphs. We believe CORE-KG lays a strong foundation for future work in automated analysis of human smuggling networks, including group discovery, entity role identification within groups, temporal graph evolution, and event prediction from legal texts and other complex narrative sources.

\section*{Acknowledgment}
This material is based upon work supported by the U.S. Department of Homeland Security under Grant Award Number 17STCIN00001-08-00. 
Disclaimer: The views and conclusions contained in this document are those of the authors and should not be interpreted as necessarily representing the official policies, either expressed or implied, of the U.S. Department of Homeland Security. 


\bibliographystyle{ACM-Reference-Format}
\bibliography{bibliography}

\appendix

\section{Per-Case Duplication and Noise Metrics}
\label{sec:appendix_percase_graphs}

\begin{figure}[h]
    \centering
    \includegraphics[width=\columnwidth]{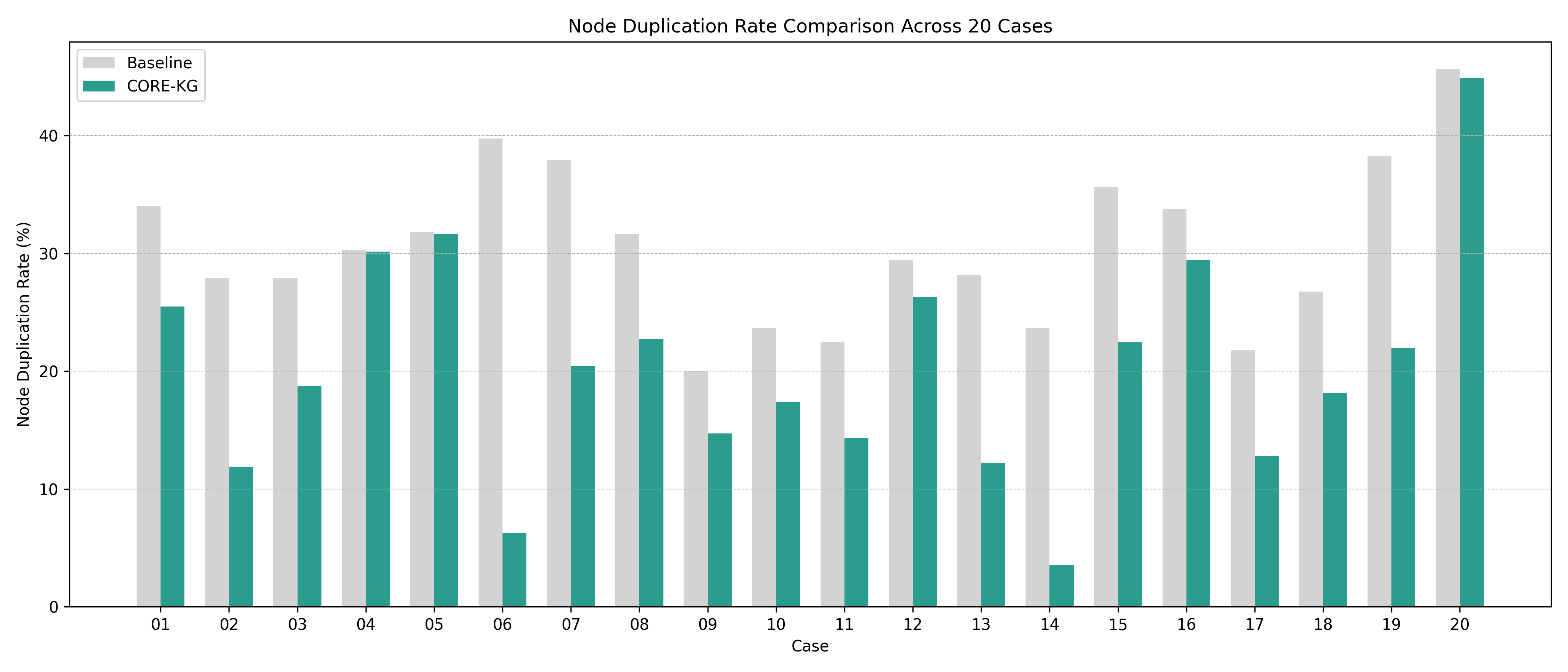}
    \caption{Node duplication rate comparison between baseline and CORE-KG across 20 legal cases.}
    \label{fig:node_duplication}
\end{figure}

\begin{figure}[h]
    \centering
    \includegraphics[width=\columnwidth]{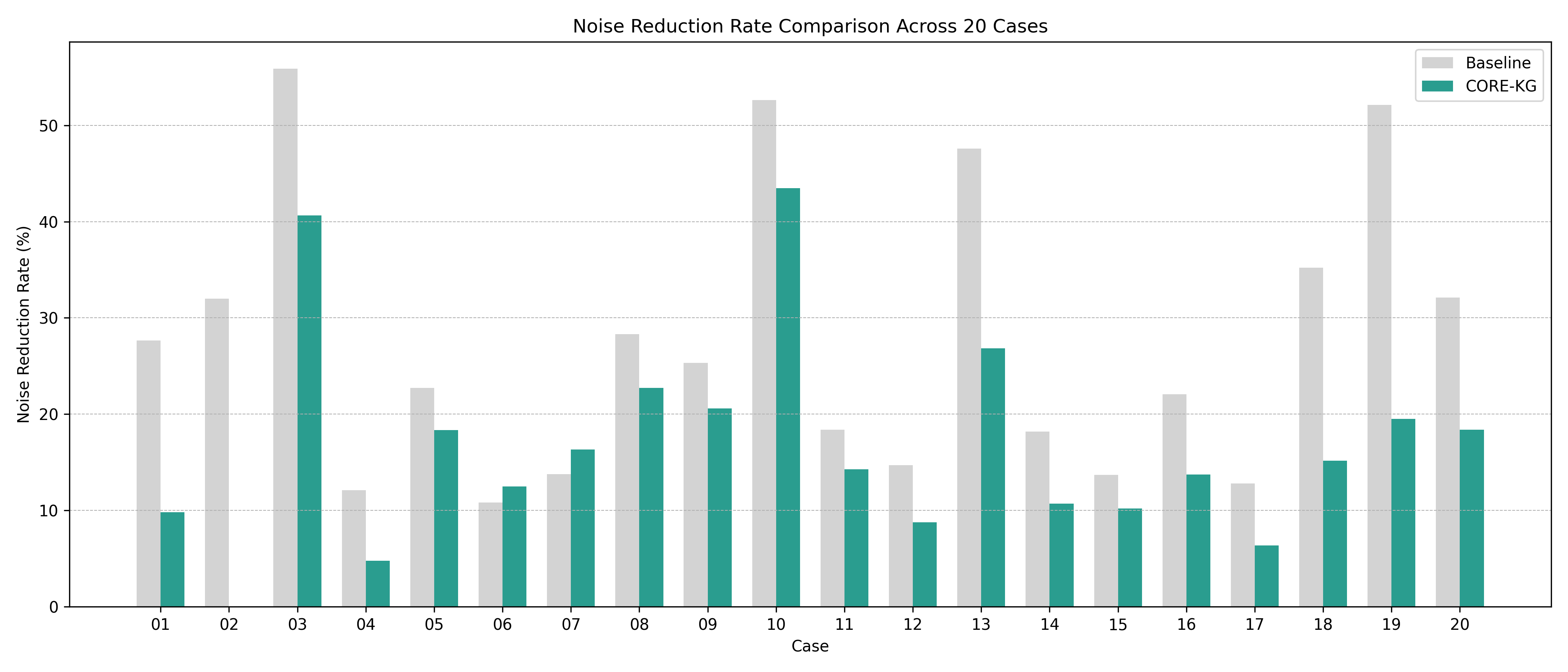}
    \caption{Noise reduction rate comparison between baseline and CORE-KG across 20 legal cases.}
    \label{fig:noise_reduction}
\end{figure}


Figures~\ref{fig:node_duplication}, and \ref{fig:noise_reduction} present the case-wise comparison of node duplication rate, and legal noise rate between the baseline GraphRAG system and our CORE-KG pipeline across all 20 legal cases.

Figure~\ref{fig:node_duplication} presents a case-wise comparison of node duplication rates between the baseline and CORE-KG across 20 legal cases. CORE-KG reduces duplication in all 20 cases, often with substantial improvements. Notably, Cases 2, 6, and 14 exhibit the most significant absolute reductions—approximately 16\%, 33.51\%, and 20\%, respectively—demonstrating CORE-KG’s effectiveness in resolving entities with high surface variability. Case 06 shows the largest improvement, reducing the duplication rate from 39.76\% to 6.25\%, due to successful consolidation of highly variable entity mentions. Case 20, while still exhibiting high duplication for both methods (45.68\% baseline vs. 44.90\% CORE-KG), remains the most challenging, reflecting limitations under extreme entity density and naming ambiguity. In a few cases, CORE-KG’s duplication rate appears close to the baseline—this occurs when the total number of entities and duplicate nodes is significantly reduced in the CORE-KG output. When the absolute number of duplicates is already low, even a small number of remaining duplicates can result in a higher relative percentage. Overall, CORE-KG achieves a relative reduction of 33.58\%, lowering the average duplication rate from 30.38\% to 20.27\%.

Figure~\ref{fig:noise_reduction} presents a comparison of legal noise rates across 20 cases. CORE-KG consistently achieves lower noise rates than the baseline, with an average reduction from 27.41\% to 16.89\%, yielding a relative improvement of 38.37\%. The largest gains are observed in Cases 2, 13, and 19, where the baseline graphs contain heavy legal boilerplate—such as court procedures, appeal details, and legal standards—that CORE-KG successfully suppresses. For example, in Case 2, noise drops from 32.00\% to 0.00\%, while Case 13 shows a reduction from 47.57\% to 26.83\%, and Case 19 from 52.13\% to 19.51\%. In fact, for Case 2, the noise reduction is particularly sharp, bringing it down to zero.
There are two cases where the performance of the baseline was better than CORE-KG—specifically, Cases 6 and 7. This is likely due to the limited presence of extractable procedural entities in these cases, which reduces the impact of CORE-KG’s in-prompt filtering mechanism. Additionally, because CORE-KG significantly reduces the total number of extracted entities through aggressive consolidation and filtering, even a small number of residual noisy nodes can result in a proportionally higher noise percentage. This effect is particularly visible when the overall entity count is low, leading to skewed ratios despite CORE-KG’s improved absolute performance.


Overall, these case-wise breakdowns reinforce the robustness of CORE-KG. While Table~\ref{tab:corekg_full_metrics} summarizes the average gains, these figures demonstrate that CORE-KG achieves consistent structural improvements across varied legal case narratives, and noise without sacrificing contextual accuracy.

\section{Entity and Relationship Extraction Prompt}

Figure~\ref{fig:llm_prompt} presents the full prompt used in our pipeline to guide the large language model (LLM) for structured extraction of entities and relationships from legal case documents. This prompt corresponds to the description in Section~\ref{sec:prompt_tuning} and reflects several task-specific strategies designed to reduce noise and improve precision in knowledge graph construction.

The prompt explicitly defines the seven targeted entity types relevant to human smuggling analysis and provides extraction instructions in a fixed order to reduce attention diffusion. It also incorporates filtering instructions to exclude high-frequency legal or government-related entities that are not structurally important to smuggling networks. In addition, each extracted relationship is assigned a confidence score based on contextual clarity, and a structured output format ensures consistent parsing.

Few-shot examples are included at the end of the prompt to illustrate the expected input-output structure. These examples help reinforce output formatting, relationship modeling, and entity-type mapping.

This prompt serves as a core component of our extraction pipeline, and its design contributes directly to the improvements reported in Section~\ref{sec:result_entity_prompt_impact}.

\begin{figure}[!htbp]
\centering
\begin{tcolorbox}[
  title=LLM Prompt for Entity and Relationship Extraction,
  colback=gray!5!white,
  colframe=gray!75!black,
  fonttitle=\bfseries,
  boxsep=0pt,         
  left=3pt,           
  right=1pt,          
  toptitle=1mm,       
  bottomtitle=1mm     
]

\tiny
\textbf{-Goal-} \\
You are an expert in Named Entity and Relationship Extraction (NER-RE) for legal case documents related to human smuggling. Your task is to extract only the specified entity types [{entity\_types}] and explicit relationships between them, without inference or completion. These outputs will be used to construct a Knowledge Graph for analyzing smuggling networks. You will receive entity definitions, input text, and structured examples—study them carefully before extraction to ensure strict factual accuracy. 

Do not extract entities corresponding to governmental organizations or entities closely related to the trial, criminal law and law procedures (e.g., jury, government, court, prosecution, etc.). These are out of scope.

\vspace{1mm}
\textbf{-Entity Type Definitions-} \\
1. \textbf{PERSON}: Any individual's name, including smugglers, agents, and undocumented migrants. \\
2. \textbf{LOCATION}: Geographical areas (e.g., city, state, country). \\
3. \textbf{ORGANIZATION}: Smuggling rings, drug cartels, and other formal groups. \\
4. \textbf{MEANS\_OF\_TRANSPORTATION}: Vehicles like car, truck, 18-wheeler. \\
5. \textbf{MEANS\_OF\_COMMUNICATION}: Tools like phone, WhatsApp. \\
6. \textbf{ROUTES}: Roads, highways, or freeways used in smuggling. \\
7. \textbf{SMUGGLED\_ITEMS}: Goods like drugs, weapons, or undocumented aliens.

\vspace{1mm}
\textbf{-Steps-}

\begin{enumerate}[left=0pt]
    \item \textbf{Entity Extraction:} Extract only explicitly stated entities of type \texttt{[{entity\_types}]}. Do not infer or complete missing information. For each, extract the following fields:
    \noindent\texttt{entity\_name} — Capitalized name as it appears.\\
    \texttt{entity\_type} — One of: \texttt{[{entity\_types}]}\\
    \texttt{entity\_description} — Detailed description of the entity’s role or attributes.

    \vspace{1mm}
    Do not extract any entities related to government or legal proceedings (e.g., court, jury, prosecution, law enforcement, etc.).

    \vspace{1mm}
    Extract entity types in the following order:
    
    \noindent\textbf{PERSON:} If a person appears with a title (e.g., “Agent R.”), extract only the full name (e.g., “R.”) as the \texttt{entity\_name} and include the title in \texttt{entity\_description}.\\
    \textbf{LOCATION:} Combine city and state into a single entity (e.g., \texttt{Laredo, Texas}).\\
    \textbf{MEANS\_OF\_TRANSPORTATION, MEANS\_OF\_COMMUNICATION, ROUTES, SMUGGLED\_ITEMS, ORGANIZATION:} Extract as relevant.

    \vspace{1mm}
    Format each entity as: \\
    \texttt{\detokenize{("entity"{tuple_delimiter}entity_name{tuple_delimiter}
    entity_type{tuple_delimiter}entity_description")}}

    \item \textbf{Relationship Extraction:} From the entities identified, extract all clearly stated relationships, even if embedded in complex sentences.

    For each relationship, extract:
    
    \noindent\texttt{source\_entity} — Source entity from step 1\\
    \texttt{target\_entity} — Target entity from step 1\\
    \texttt{relationship\_description} — Explanation of the connection\\
    \texttt{relationship\_strength} — Score between 0–10: 
    \begin{itemize}
        \item \textbf{0–3 (Weak):} Indirect or uncertain (e.g., “may have…”)
        \item \textbf{4–6 (Moderate):} Explicit but lacks strong context
        \item \textbf{7–10 (Strong):} Clear, direct, and contextually supported
    \end{itemize}

    \vspace{1mm}
    Format each relationship as: \\
    \texttt{("relationship"\{tuple\_delimiter\}source\_entity\{tuple\_delimiter\}
    target\_entity\{tuple\_delimiter\}relationship\_description\{tuple\_delimiter\}
    relationship\_strength")}

    \item \textbf{Filter Government Entities:} If any government-related entities or relationships are mistakenly extracted, remove them.

    \item \textbf{Output Format:} Return all extracted entities and relationships as a single list using \texttt{\{record\_delimiter\}} as the separator.

    \item \textbf{Completion Token:} End the output with: \texttt{\{completion\_delimiter\}}
\end{enumerate}

\vspace{1mm}
\textbf{- Few-shot Examples-} \\
\textbf{Example 1} \\
\textit{Input:} \texttt{Smugglers from the Horizon Smuggling Ring used WhatsApp.} \\
\textit{Output:} \\
\noindent\texttt{("entity"\{tuple\_delimiter\}SMUGGLERS\{tuple\_delimiter\}PERSON\{tuple\_delimiter\}...)} \{record\_delimiter\} \\
\texttt{("entity"\{tuple\_delimiter\}WHATSAPP\{tuple\_delimiter\}MEANS\_OF\_COMMUNICATION\{tuple\_delimiter\}...)} \{record\_delimiter\} \\
\texttt{("relationship"\{tuple\_delimiter\}SMUGGLERS\{tuple\_delimiter\}WHATSAPP\{tuple\_delimiter\}...)} \{record\_delimiter\} \\
\texttt{\{completion\_delimiter\}} \\
...

\end{tcolorbox}
\caption{Prompt used for entity and relationship extraction from case documents.}
\label{fig:llm_prompt}
\end{figure}

\section{Comparison of Graph Quality for a Representative Legal Case}

Figures~\ref{fig:corekg_single_graph} and~\ref{fig:grag_single_graph} present a side-by-side comparison of the knowledge graphs generated by the baseline GraphRAG system and our proposed CORE-KG pipeline for the same input case 2.

To improve interpretability, relationship labels have been omitted and node sizes are scaled by their degree (i.e., the number of connections each entity participates in), which serves as a visual proxy for node centrality. 

Figure~\ref{fig:grag_single_graph} illustrates the baseline knowledge graph generated using GraphRAG for Case 2. The graph contains several redundant nodes, including surface-level variants of the same entity, highlighted using colored rectangles. The same color is used to indicate duplicate nodes that refer to the same entity expressed in different forms. The baseline system redundantly represents the main actor, \texttt{A.Y.}, as multiple separate nodes such as \texttt{Y.} and \texttt{A.Y.}. These entities appear disconnected or only indirectly linked, fragmenting the narrative and inflating the graph’s structural complexity. Additionally, legal and procedural terms such as \texttt{Court}, \texttt{Defense}, and \texttt{Court of Appeals} dominate the graph, contributing to legal noise despite being irrelevant to the underlying smuggling network structure.  

In contrast, Figure~\ref{fig:corekg_single_graph} shows CORE-KG correctly merging the coreferent mentions of the main actor into a single \texttt{A.Y.} node. Other coreferences have also been resolved and resultant into clean and more coherent graph. This result into reduction in the total node count from 86 in baseline graph to 42 in core-kg graph. Moreover the relationship count has reduced from 117 in the baseline graph to the 70 in the CORE-KG  graph. The result is a cleaner, more coherent structure that better represents the underlying smuggling narrative. Government-related entities are filtered out during prompt-guided extraction, reducing noise and improving analytical clarity.

These visualizations support the quantitative results in Table~\ref{tab:corekg_full_metrics} and the qualitative analysis in Section~\ref{sec:result_entity_prompt_impact}, demonstrating that CORE-KG produces graphs that are both semantically aligned and structurally faithful to the case narrative.

\begin{figure*}[t]
\centering
\includegraphics[width=0.9\linewidth]{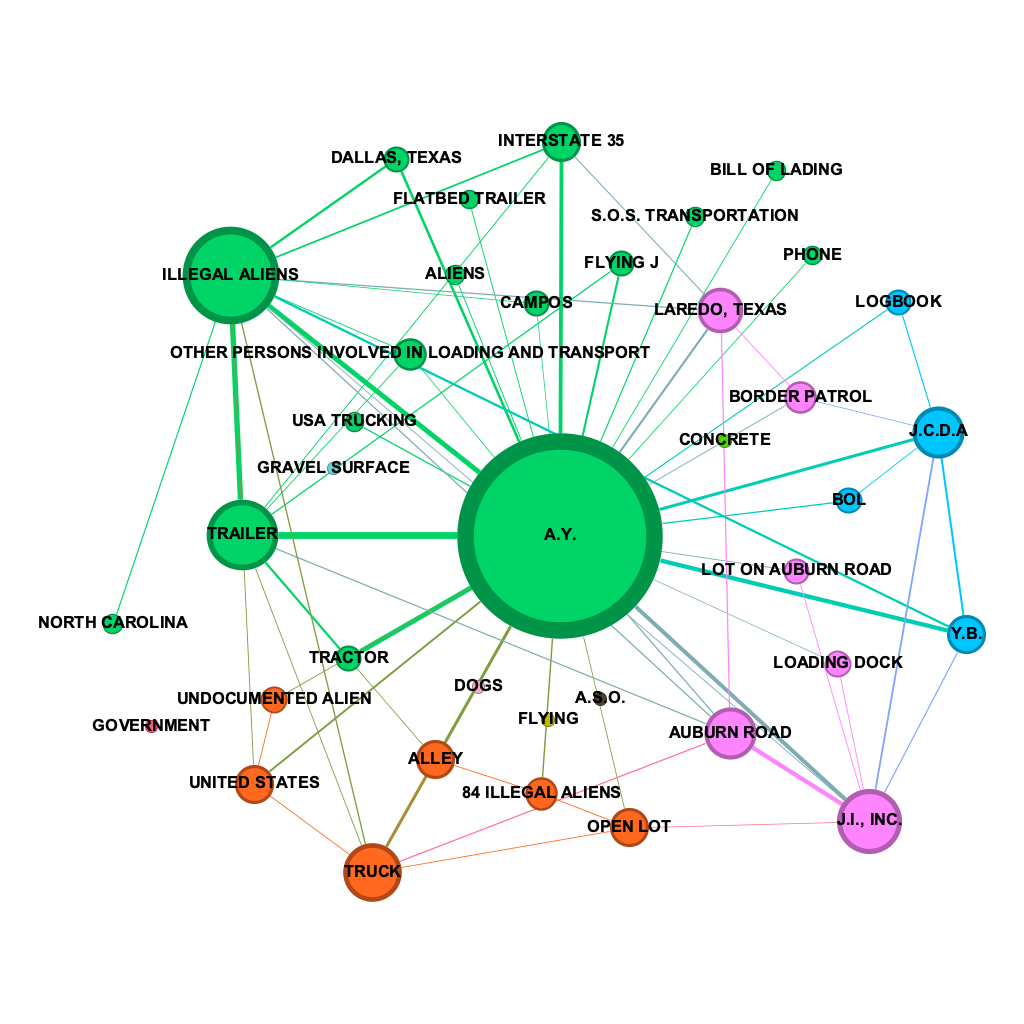}
\caption{Knowledge graph generated by CORE-KG for a representative legal case. The graph demonstrates resolved coreference, improved coherence, reduced legal noise, and more precise entity linking compared to the baseline output.}
\label{fig:corekg_single_graph}
\end{figure*}

\begin{figure*}[t]
\centering
\includegraphics[width=0.9\linewidth]{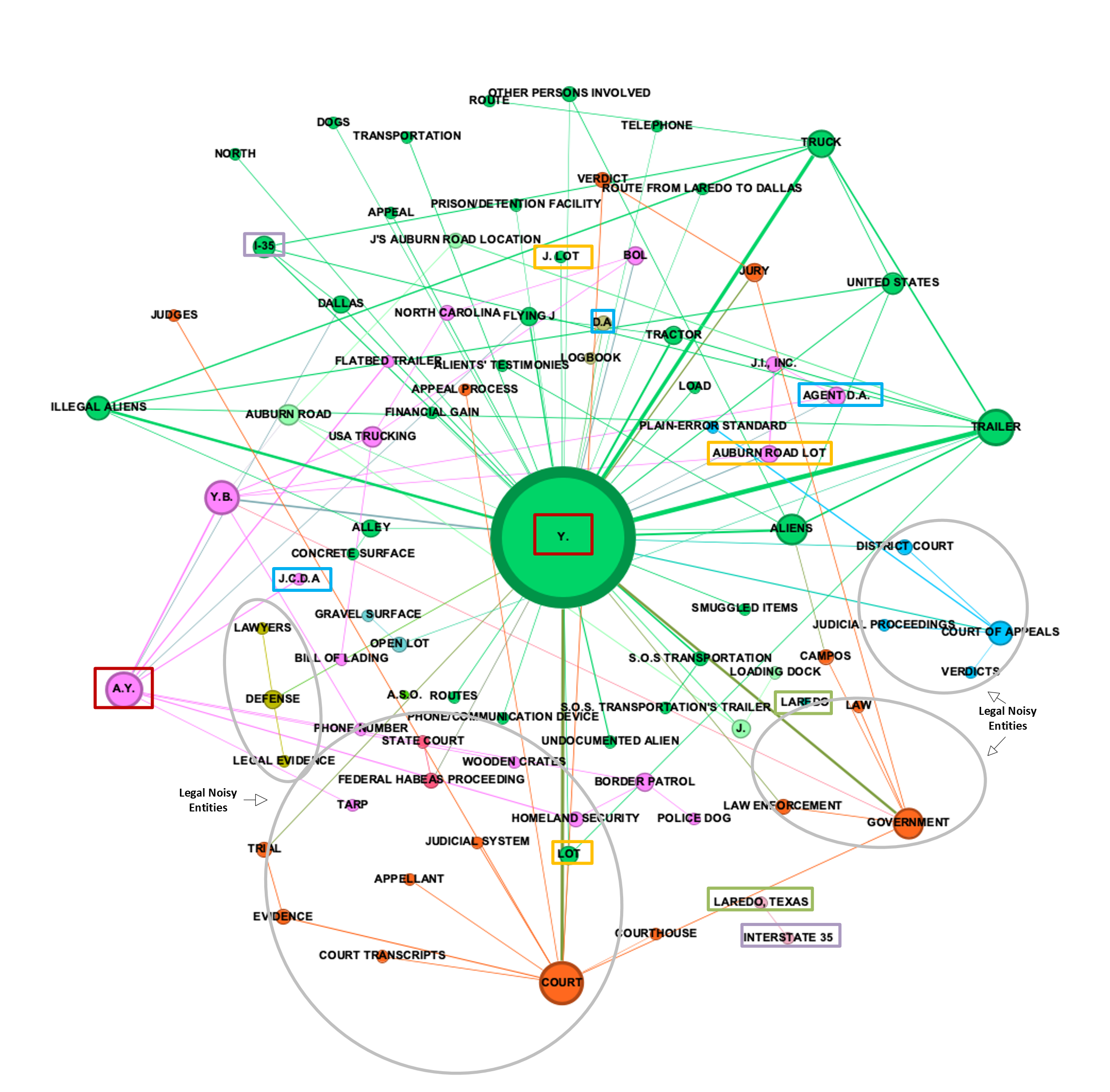}
\caption{Baseline knowledge graph generated using GraphRAG for a representative legal case. The graph contains several redundant nodes, generic entities, and visually dense connections, which reduce overall clarity. Key duplicate nodes are highlighted using rectangles of the same color, indicating repeated entities that fragment the graph structure.}
\label{fig:grag_single_graph}
\end{figure*}

\section{Coreference Resolution Prompt — Person Entity Type}
\label{appendix:coref_prompt_person}

Appendix Figure~\ref{fig:coref_prompt_person} displays the full prompt used for resolving coreferences related to the \texttt{Person} entity type. This prompt is a central component of our type-aware coreference resolution strategy, described in Section~\ref{sec:coref_resolution_prompt_design}.

The design reflects our structured approach, with components including a persona definition, task descripton, contextual information, entity-type-specific resolution rules, and few-shot examples tailored to the legal domain. The prompt emphasizes strict text preservation, title removal, and disambiguation logic when multiple persons share surnames. It also includes guidance for resolving both singular and plural references like “the defendant” and “the defendants.” The few-shot examples at the end of the prompt illustrate how the system should resolve references across varying contexts—capturing patterns such as title abbreviation (e.g., BPA S.), full title references (e.g., Border Patrol Agent H.D.I.), and compound name disambiguation. This structured design helps ensure consistency, accuracy, and generalization across diverse legal case documents in our dataset.
\vskip 2in

\begin{tcolorbox}[
  title=Prompt for Coreference Resolution — Person Entity Type,
  colback=gray!5!white,
  colframe=gray!75!black,
  fonttitle=\bfseries,
  boxsep=2pt,
  left=2pt,
  right=2pt,
  toptitle=1mm,
  bottomtitle=1mm,
  breakable,
  enhanced,
  rounded corners
]

\tiny
\textbf{- Goal -} \\
You are a highly precise and intelligent coreference resolution system designed to support named entity recognition (NER) and knowledge graph construction. Your task is to resolve all coreferences related to the Person entity type—including roles and titles (e.g., Defendant, Officer, Agent)—in a given input text, while strictly preserving its original structure and wording. The resolved output will be used for extracting person entities and relationships in the context of human smuggling networks. Therefore, maintaining accuracy and consistency is critical. Do not summarize, explain, or alter the text—only return the full, unmodified input with Person coreferences resolved according to the rules below.

Note: This is an unsupervised coreference resolution task. The instructions are designed to guide you in resolving person-related references. While examples are provided, they do not cover all scenarios. You must infer and apply coreference logic based on contextual understanding, even when phrasing or structure varies.
\vspace{1mm}
\textbf{- Coreference Resolution Rules — Person Entity Type -}

\begin{itemize}[left= 2pt]
  \item After a person is introduced with their full name (e.g., Paul Silva), replace all subsequent mentions—including last name only (e.g., S.), role + last name (e.g., Agent S.), and abbreviated forms (e.g., BPA S.)—with the full name only.
  \item In all coreference resolutions, strip titles from mentions. For example, "Agent I." or "Agent J.C.D.A." should resolve to "Hector D.I." or "J.C.D.A.".
  \item For compound names, match based on the final component (e.g., I., R.) and resolve to the full name.
  \item If two or more individuals share a last name, resolve ambiguous mentions conservatively—default to the most recently introduced full name unless context clearly indicates otherwise.
  \item If abbreviated titles appear (e.g., BPA, Agent, Officer + Last Name), remove the title and resolve to the full name.
  \item If a person is introduced as "Defendant M.D.J.G.", resolve it to "M.D.J.G" immediately and throughout.
  \item If someone is introduced as "Border Patrol Agent H.D.I", retain this in the first mention, but resolve all later mentions (e.g., "Agent I.") to "H.D.I".
  \item Apply all replacements across the entire document, including headers, transcripts, footnotes, and end-of-document text.
\end{itemize}

\textbf{Multiple Defendants:}
\begin{itemize}[left= 2pt]
  \item If multiple defendants are introduced, resolve "the defendants" to a comma-separated list of their full names, in the order introduced.
  \item "The defendant" (singular) should resolve to the most recently mentioned full defendant name unless context indicates otherwise.
  \item Always resolve all such role-based mentions, even in peripheral document sections.
\end{itemize}

\vspace{1mm}
\textbf{- Examples -}

\textbf{Example 1:} \\
Input: Border Patrol Agent B.S. observed the vehicle. BPA S. contacted another agent. \\
Output: Border Patrol Agent B.S. observed the vehicle. B.S. contacted another agent.

\textbf{Example 2:} \\
Input: Border Patrol Agent H.D.I led the operation. I. coordinated with the local sheriff. \\
Output: Border Patrol Agent H.D.I. led the operation. H.D.I coordinated with the local sheriff.
....

\vspace{1mm}
\textbf{- Input Text -} \\
Resolve all Person entity coreferences in the following document, including those in footnotes and headers. Return only the modified text. If none exist, return the input unchanged. Do not summarize or explain.
\texttt{Input\_text: \{input\_text\}} \\
\texttt{Output:}

\end{tcolorbox}

\vspace{2mm}
\captionsetup{type=figure}
\captionof{figure}{Prompt used for resolving coreferences related to Person entities. This includes resolution rules, behavioral constraints, and illustrative examples.}
\label{fig:coref_prompt_person}

\end{document}